\documentclass[letterpaper, 10 pt, conference]{ieeeconf}  

\IEEEoverridecommandlockouts                              
\usepackage{cite}
\usepackage[cmex10]{amsmath}
\usepackage{algorithmic}
\usepackage{array}
\usepackage{float}
\usepackage{graphics} 
\usepackage{epsfig} 
\usepackage{mathptmx} 
\usepackage{times} 
\usepackage{amsmath} 
\usepackage{amssymb}  
\usepackage{dblfloatfix}
\usepackage[hyphens]{url}
\usepackage{subcaption}
\usepackage{lipsum}
\usepackage{adjustbox,lipsum}
\usepackage{booktabs}
\usepackage{calrsfs}
\DeclareMathAlphabet{\pazocal}{OMS}{zplm}{m}{n}
\usepackage{multirow}
\usepackage{diagbox}
\usepackage{xcolor}
\usepackage{tabularx,ragged2e,booktabs,caption}
\usepackage[british]{babel}
\usepackage{textcomp}
\usepackage{hyphenat}
\usepackage{hyperref}
\usepackage[font=small]{caption}
\usepackage{tikz}
\def\checkmark{\tikz\fill[scale=0.3](0,.35) -- (.25,0) -- (1,.7) -- (.25,.15) -- cycle;} \usepackage{balance}

\begin{document}

\title{\LARGE \bf Orientation Attentive Robotic Grasp Synthesis with\\Augmented Grasp Map Representation}

\author{Georgia Chalvatzaki$^{1^{*}}$, Nikolaos Gkanatsios$^{2^{*,\dagger}}$, Petros Maragos$^3$ and Jan Peters$^{1}$\\
\thanks{This project received funding from the RoboTrust and the Skills4Robots projects. Experimental computations have been conducted on an Nvidia DGX-1 at TU Darmstadt.}
\thanks{$\dagger$ Work completed while Nikolaos Gkanatsios was at National Technical University of Athens.}
\thanks{$^1$Intelligent Autonomous Systems, Technische Universit\"{a}t Darmstadt, Hochschulstr. 10, 64289 Darmstadt, Germany \tt \small  georgia@robot-learning.de, \tt \small mail@jan-peters.net}
\thanks{$^2$Robotics Institute, Carnegie Mellon University, 5000 Forbes Ave., 15213 Pittsburgh, PA, USA \tt \small ngkanats@andrew.cmu.edu}
\thanks{$^3$School of E.C.E., National Technical University of Athens, 15773, Athens, Greece \tt \small maragos@cs.ntua.gr}
\thanks{*Authors contributed equally}
}

\maketitle
\thispagestyle{empty}
\pagestyle{empty}

\begin{abstract}
Inherent morphological characteristics in objects may offer a wide range of plausible grasping orientations that obfuscates the visual learning of robotic grasping. Existing grasp generation approaches are cursed to construct discontinuous grasp maps by aggregating annotations for drastically different orientations per grasping point. Moreover, current methods generate grasp candidates across a single direction in the robot's viewpoint, ignoring its feasibility constraints.
In this paper, we propose a novel augmented grasp map representation, suitable for pixel-wise synthesis, that locally disentangles grasping orientations by partitioning the angle space into multiple bins. Furthermore, we introduce the ORientation AtteNtive Grasp synthEsis (ORANGE) framework, that jointly addresses classification into orientation bins and angle-value regression. The bin-wise orientation maps further serve as an attention mechanism for areas with higher \textit{graspability}, i.e. probability of being an actual grasp point. We report new state-of-the-art 94.71\% performance on Jacquard, with a simple U-Net using only depth images, outperforming even multi-modal approaches. Subsequent qualitative results with a real bi-manual robot validate ORANGE's effectiveness in generating grasps for multiple orientations, hence allowing planning grasps that are feasible. Code is available at \url{https://github.com/nickgkan/orange}.

\end{abstract}

\section{Introduction}
\begin{figure}
    \includegraphics[width=0.5\textwidth]{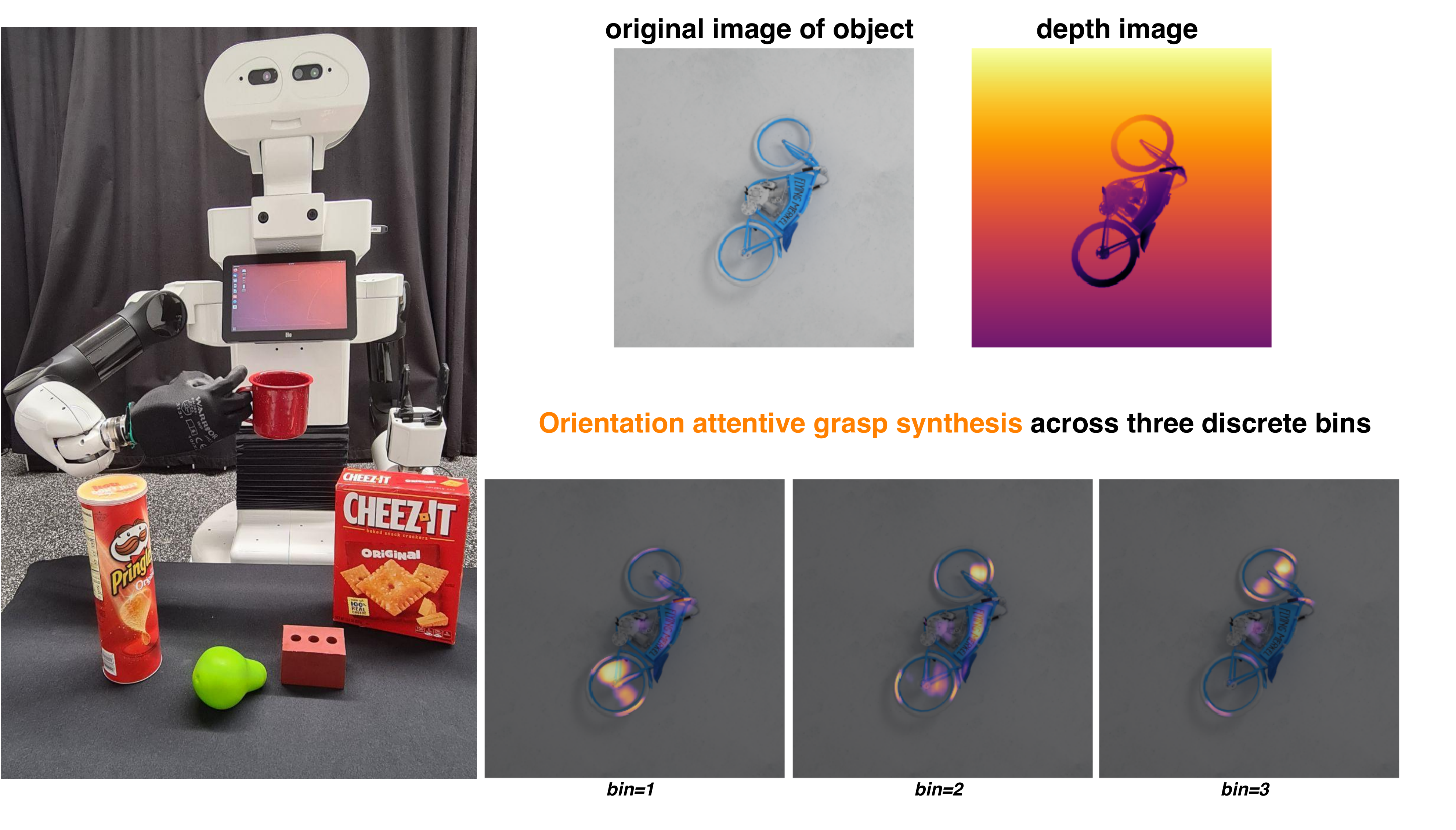}
  \caption{\small \textbf{Left:} Everyday objects have variable morphologies making robotic grasping challenging. TIAGo++ \cite{pages2016tiago} is equipped with an underactuated hand and a gripper, hence is able to perform different types of grasps. \textbf{Right:} Our \textit{ORANGE} framework effectively detects grasping points of different orientations, thanks to its augmented grasp map representation and the orientation attention that focuses on points with maximum graspability.}
    \label{fig:robot_motivation}
    \vspace{-0.82cm}
\end{figure}

Successful robotic grasping in unstructured environments (Fig. \ref{fig:robot_motivation}) is a long aspiration towards the successful migration of robots into human-inhabited environments. However, this includes several sub-problems to be solved, like perception, grasp planning, and control. For that reason, grasping objects of different shapes, textures and sizes \cite{weng2020multi, cui2020grasp, wu2019learning,manuelli2019kpam}, has been explored both in an analytical \cite{Shimoga96robotgrasp,SAHBANI2012} and data-driven fashion \cite{Bohg2013,shao2020unigrasp, mahler2019learning}. In general, the notion of a grasp can be parameterized by a point on the object, defining the \textit{approaching vector} according to which the robot's end-effector center should align, a 3-D angle with which the robot's tool should approach the grasp point and the initial configuration of the tool w.r.t. to its optimal width for performing the grasp \cite{Ekvall2007}.  

The advantages in Deep Learning (DL), along with the introduction of low-cost RGB-D sensors and the creation of large datasets, gave an increasing benefit to data-driven approaches for robotics perception. During the last years, such datasets have arisen for robot grasping \cite{Lenz2015,Jacquard,YCB,fang2019graspnet,Fang_2020_CVPR}, containing a multitude of graspable objects usually found in households, that are suitable for robotic hands and grippers. In this light, \textit{grasp rectangles} became the typical representation usually employed in cascaded deep networks for detecting possible grasp candidates \cite{Lenz2015, Wang2016,Redmon2015, Kumra2017, Watson2017,Guo2017}. Several approaches try to transfer successful computer vision methods, like object detection with bounding box regression \cite{NIPS2015Faster}, for detecting antipodal grasps on objects from RGB data. These approaches predict and rank thousands of grasp candidates \cite{Zhou2018,Zhang,Zhang2,Chu2018,cheng2020high}, requiring many computational resources, while they are limited to static environments and precise camera calibration.

Recently, pixel-wise approaches attempted to confront these issues \cite{Johns2016,morrison2018closing,mahler2017dex,zeng2018learning,satish2019policy, cai2019metagrasp,GGCNN_IJRR} exploring the generative capabilities of a convolutional neural network (CNN), in order to estimate grasps for real-time performance in robotics.
The continuous orientation estimation is crucial \cite{morrison2019multi,lin2020using,choi2020hierarchical}, especially for reactive grasp planning, e.g. in cases when the camera is mounted on the robot (Fig. \ref{fig:robot_motivation}) and changes its perspective. While several approaches rely on reinforcement learning of grasping policies from visual data \cite{mahler2017dex,fang2020learning, satish2019policy, song2020grasping}, those are usually collapsing into learning a single grasping distribution mode. Potentially, we require a grasp generator to provide multiple possible candidates \cite{murali20206}, so that we will be able to plan robot actions across multiple directions in the robot's viewpoint and feasible workspace.
Intuitively, humans argue about the object's shape and navigate their hand with appropriate orientation and opening in order to perform the grasp. However, existing supervised vision-based learning approaches suffer from ambiguities due to multiple overlapping grasping boxes with different orientations, failing to accurately represent possible grasps. 

Among the different issues that hinder efficient and generalized robot grasping across various platforms and end-effectors, in this paper, we tackle the problem of disentangling the possible orientations per grasp point. To this end, we propose a novel augmented grasp representation that parses annotated grasps into multiple orientations bins. Stemming from this representation, we introduce an \textit{orientation-attentive} method for predicting pixel-wise grasp configurations from depth images. We classify the grasps according to their orientations into discrete bins, while we regress their values for continuous estimation of the grasp orientation per bin. Moreover, this orientation map acts as a bin-wise \textit{attention mechanism} \cite{NIPS2017_7181} over the grasp quality map, to teach a CNN-based model to focus its attention on the actual grasp points of the object. The proposed method, named \textit{ORANGE} (ORientation AtteNtive Grasp synthEsis), is model-agnostic, as it can be interleaved with any CNN-based approach capable of performing segmentation while boosting their performance for improved grasp predictions. 

ORANGE achieves state-of-the-art results on the most challenging grasping dataset \cite{Jacquard}, acquiring 94.71\% using only the depth modality, against all other related methods. Knowledge from ORANGE can also be easily transferred and leads to significantly accurate predictions on the much smaller dataset Cornell \cite{Lenz2015}. Moreover, our analysis is supported by robotic experiments, both in simulation and with a real robot (Fig. \ref{fig:robot_motivation}). Our physical experiments show the importance of disentangling the grasp orientation for achieving an efficient robot grasp planning, while also highlighting other parameters that affect the grasp success.

\begin{figure*}
\vspace{-0.1cm}
\centering
\begin{subfigure}{.65\textwidth}
  \centering
  \includegraphics[width=.95\linewidth]{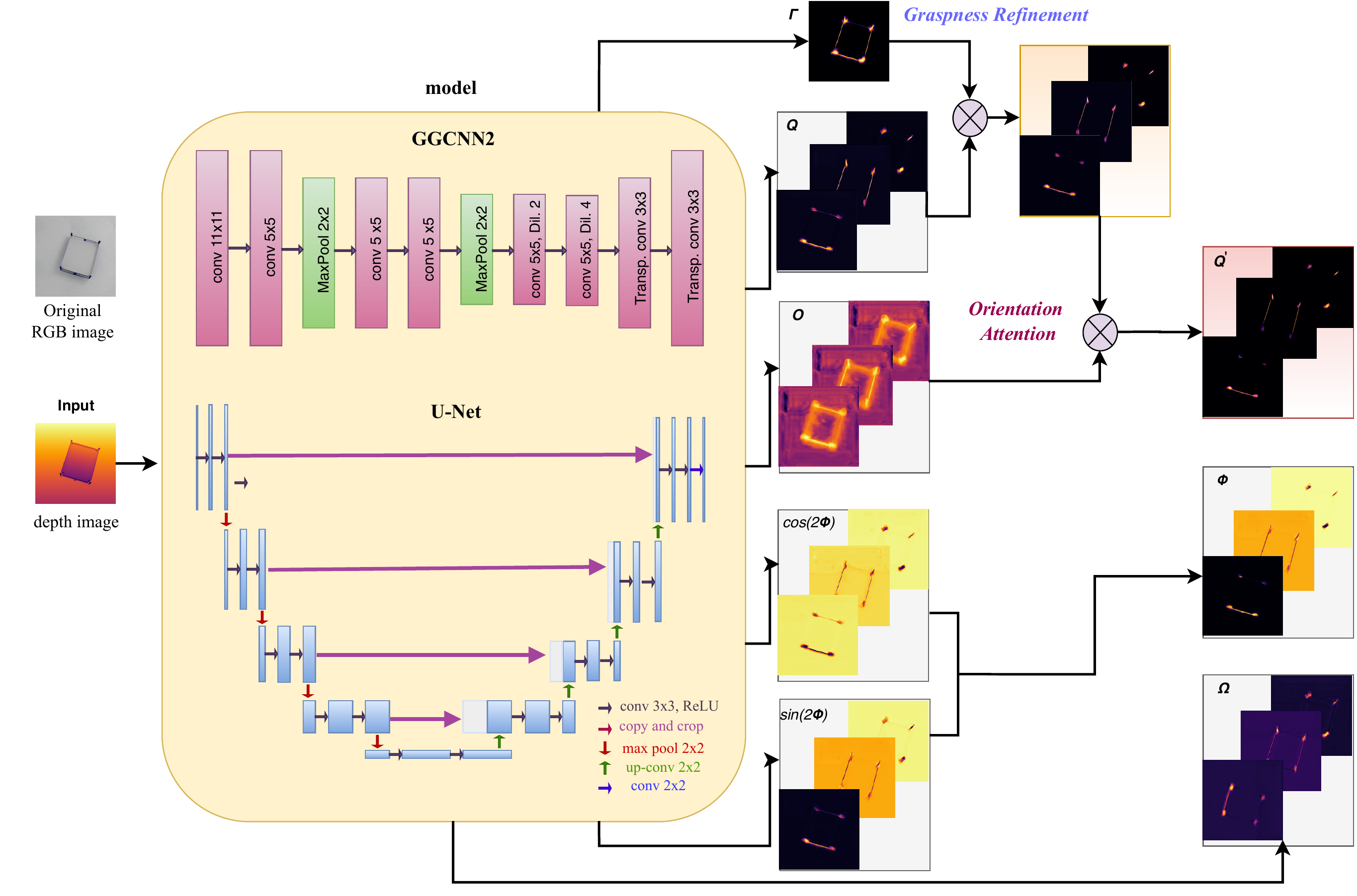}
  \caption{}
  \label{fig:framework}
\end{subfigure}%
\begin{subfigure}{.35\textwidth}
  \centering
  \includegraphics[width=.95\linewidth]{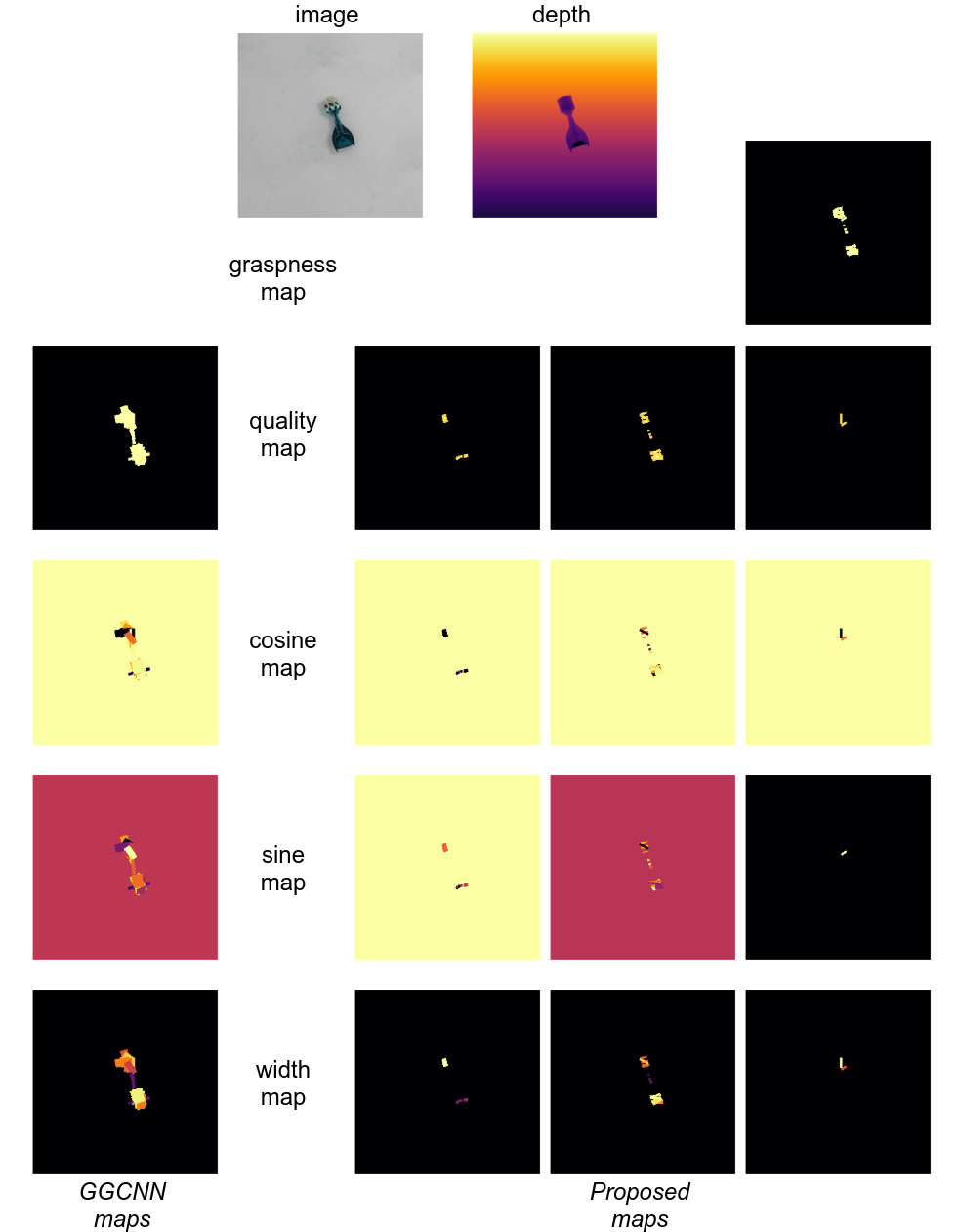}
    \vspace{-0.25cm}
  \caption{}
  \label{fig:gt_maps}
\end{subfigure}
\vspace{-0.25cm}
\caption{\textbf{(a)} Overview of the ORANGE architecture. An augmented grasp map representation, that fuses continuous and discrete information, drives the transformation of the depth image into a set of grasping boxes. The discretized orientation map serves as an attention force that focuses on local maxima of graspability. Please refer to Sec.~\ref{sec:method} for a thorough explanation of the symbols used in this figure. \textbf{(b)} Comparison of the target representation for GGCNN2 (left column) and the proposed 3-bin method (right 3 columns). GGCNN2 maps suffer from highly overlapping boxes that lead to discontinuities, while their binary quality map is a dense region that lies further than the object's boundaries. Contrary to that, our maps are sparse and clear from overlaps, while the quality maps contain rigid areas with a well-defined maximum. Our ``graspability'' map roughly approximates the object's segmentation mask.}
\label{fig:main}
\vspace{-0.4cm}
\end{figure*}

\section{Problem Statement} \label{sec:problem_statement}
Grasp synthesis refers to finding the optimal grasp configuration $\bf{g}=\{x,y,z,\phi,w,q\}$, containing the grasp center $\{x,y,z\}$ to which the robotic hand should be aligned, the orientation $\phi$ around the $z$ axis and the required fingers' or jaws' opening (width) $w$. A quality measure $q$ characterizes the success of the respective grasp configuration. For a (depth) image $\mathbf{I}$, grasp synthesis is the problem of finding the \textit{grasp map} \cite{GGCNN_IJRR}:
\begin{equation}\label{eq:grasp_map1}
\mathbf{G}=\{{\Phi, \Omega, Q}\} \in \mathbb{R}^{3 \times H \times W}
\end{equation}
where $\Phi, \Omega,Q$ are each of them a map in $\mathbb{R}^{H \times W}$, containing the pixel-wise values of $\phi, w, q$
respectively. $\mathbf{G}$ can be approximated through a learnt mapping $\mathbf{I} \xrightarrow{\hat f_{\theta}} \mathbf{G}$ using a deep neural network ($\theta$ being its weights). The best visible grasp configuration can now be estimated as $\bf{\bar g^{*}}=\mathrm{arg}\max\limits_{Q} \mathbf{G}$.

\section{Method} \label{sec:method}
The grasp maps constructed by current pixel-wise learning approaches \cite{GGCNN_IJRR,song2019deep,Wang,kumra2019antipodal} are prone to discontinuities that cause performance to saturate, due to the overlapping grasping orientations per point. Motivated by the need of acquiring approaching grasp vectors from multiple orientations, we introduce an augmented grasp map representation, that fuels both the continuous orientation estimation, commonly treated as a regression problem, and a discrete classification (Fig. \ref{fig:framework}). In the following, we first present an extensive analysis on the complexity of grasp datasets like Jacquard, as well as previous approaches on grasp map estimation, then we discuss our specific design choices for tackling the problem of visual robotic grasp synthesis from depth images.

\begin{figure}
        \centering
    \begin{adjustbox}{max width=0.5\textwidth}
{
    \begin{tabular}{c|c|c|c}
         Parameter & GGCNN2 \cite{GGCNN_IJRR} & Our method & Benefit \\ \hline
         Discretize angle &            & \checkmark & Less overlaps \\ \hline
         Map dimensions   & 2-d        & 3-d        & Less overlaps \\ \hline
         Quality map      & Binary     & Non-binary & Accurate centering \\ \hline
         Postprocessing   & Gaussian filtering & None & Faster \\ \hline
         Picked jaws' size & All        & Minimum    & Improved segmentation \\ \hline
         Handle ovelaps  & Overwrite  & Keep minimum & Rigid map spaces \\ \hline
         Max. IoU@0.25   & 96.24      & 97.32 & Better reconstruction \\ \hline
         Max. IoU@0.30   & 94.96      & 95.83 & Better reconstruction \\ \hline
         Max. IoU@0.50   & 84.72      & 89.38 & Better reconstruction \\ \hline
   \end{tabular}
   }
\end{adjustbox}
    \captionof{table}{\small Comparison of design choices between the proposed method and prior literature \cite{GGCNN_IJRR} concerning grasp map construction. Our real-valued maps resolve ambiguities due to overlaps, leading to better reconstruction ability.}
    \label{tab:map_comparison}
    \vspace{-0.65cm}
\end{figure}
\subsection{Revisiting Grasp Map Representation} \label{sec:dataset_discussion}
Jacquard is currently one of the most diverse and densely annotated grasping datasets with 54000 images and 1.1 million grasp annotations. Grasps are represented as rectangles with given center, angle, width (gripper's opening) and height (jaws' size). The annotations are simulated and not human-labeled, resulting into multiple overlapping boxes considering all possible grasp orientations per grasp point and many different jaw sizes. Box annotations are invariant to the jaws’ size, leaving it as a free variable to be arbitrarily chosen during evaluation.

The authors of \cite{GGCNN_IJRR} proposed a grasp map representation, generating pixel-wise quality, angle and width maps, by iterating over the annotated boxes and stacking binary maps, equal to the value of interest inside the box and zero elsewhere. Since the quality map is binary, it is indifferent to the order of the boxes and equivalent to iterating only on the boxes with the maximum jaws' size. For angle and width maps however, overlapping boxes with different centers and angles will be overwritten by the box that appears last in the list, hence leading to discontinuities. Crucially, a binary quality map does not ensure a valid maximum: all non-center points inside an annotated box are maxima as well, and have equal probability of being selected as a grasp center. Due to these facts, a hypothetical regressor that perfectly predicts the evaluation GT maps fails to reconstruct the annotated bounding boxes and scores only $\sim{96.2}\%$ using the Jaccard (Intersection over Union-IoU) \cite{Chu2018} index at the 0.25 threshold, while its performance degrades rapidly towards higher thresholds (Table~\ref{tab:map_comparison}). 

\subsection{Grasp Maps with Discretized Orientation}\label{sec:grasp_map}
To tackle the aforementioned challenges, we part from recent approaches on pixel-wise grasp synthesis \cite{GGCNN_IJRR,song2019deep,Wang} and partition the angle values into $N$ bins, so as to minimize the overlaps of multiple angles per point. Since we are dealing with antipodal grasps, it is sufficient to predict an angle in the range of $\{-\pi/2,\pi/2 \}$. We, thus, proceed to construct 3-dimensional maps of size $H \times W \times N$, where each bin corresponds to a range of $180/N$ degrees. Note, however, that we do not discretize the angles' values: we instead place them inside the corresponding bins. For the remaining overlaps, we pick the value with the smallest angle, ensuring that the network is trained on a valid GT angle value, instead of some statistics of multiple values (e.g. mean or median), while remaining invariant to the order of the annotations.

To overcome the information loss from constructing binary maps, we create soft quality maps that contain ones on the exact positions of the centers of the boxes, while their values degrade moving towards the boxes' edges (Fig.~\ref{fig:gt_maps}). We find this significant for the networks to learn to maximize the quality value on the actual grasp points, and do not acquire strong Gaussian filtering \cite{GGCNN_IJRR} and consequently reduces post-processing time. One remaining issue is the multiple instances of the same grasp centers and angles using different jaw sizes. We construct our augmented maps picking the smallest jaw size available, i.e. closer to the boundaries of the objects' shape. Intuitively, the annotated quality map gives a rough estimate of the object’s segmentation mask, which appears important for extracting grasp regions. During evaluation, we adopt the half jaw size as in \cite{GGCNN_IJRR}\footnote{See \url{https://github.com/dougsm/ggcnn}} to be directly comparable. Although having to estimate this parameter hurts performance, our approach still achieves large reconstruction ability. 

We reformulate Eq.~(\ref{eq:grasp_map1}) to consider $N$ orientation bins:
\begin{equation}\label{eq:grasp_map2}
\mathbf{G}=\{{\Phi, \Omega, Q, O, \Gamma}\} \in \mathbb{R}^{(4 \times N)+1 \times H \times W}
\end{equation}
where $\Phi \in \mathbb{R}^{N \times H \times W}$ is the angle map. For facilitating learning, we adopt the angle encoding suggested by \cite{hara2017designing,GGCNN_IJRR} into the cosine, sine components that lie in the range of $[-1, 1]$. Since the antipodal grasps are symmetrical around $\pm \frac{\pi}{2}$, we employ the sub-maps for $cos(2\Phi_i)$ and $sin(2\Phi_i)$ $\forall \Phi_i$ with $i\in N$ bins. The angle maps are then computed as: $\Phi =\frac{1}{2}\arctan \frac{sin(2\Phi)}{cos(2\Phi)}$. $\Omega \in \mathbb{R}^{N \times H \times W}$ represents the gripper's width map. $Q \in \mathbb{R}^{N \times H \times W}$, is a real-valued quality map, where `$1$' indicates a grasp point with maximum visible quality. $O \in \mathbb{R}^{N \times H \times W}$ is a binary orientation map where `$1$' indicates a filled angle bin in the respective position. $\Gamma \in \mathbb{R}^{1 \times H \times W}$ is the pixel-wise ``graspability'' map. This binary map contains `$1$s' only in the annotated grasp points of the object w.r.t. the image $\bf{I}$, and helps assessing the graspability of the pixels, i.e. the probability of representing grasp points of the real world. An example of the constructed grasping maps, as well as a comparison to those of \cite{GGCNN_IJRR} can be seen in Fig.~\ref{fig:gt_maps}.

\subsection{ORANGE: Orientation-attentive grasp synthesis}\label{sec:orange}
The proposed framework, ORANGE, depicted in Fig. \ref{fig:framework}, is model-agnostic; it suffices to employ any CNN-based model that has the capacity to segment regions of interest. Then, an initial depth image is processed to output an augmented grasp map $\mathbf{G}$, as in Eq.~(\ref{eq:grasp_map2}). $\Phi,\mbox{ } \Omega, \mbox{ } Q,\mbox{ } O$ and $\Gamma$ are combined to reconstruct the grasps centers, angles and widths.

\begin{figure}
    \vspace{-0.1cm}
    \centering
    \includegraphics[width=.95\linewidth]{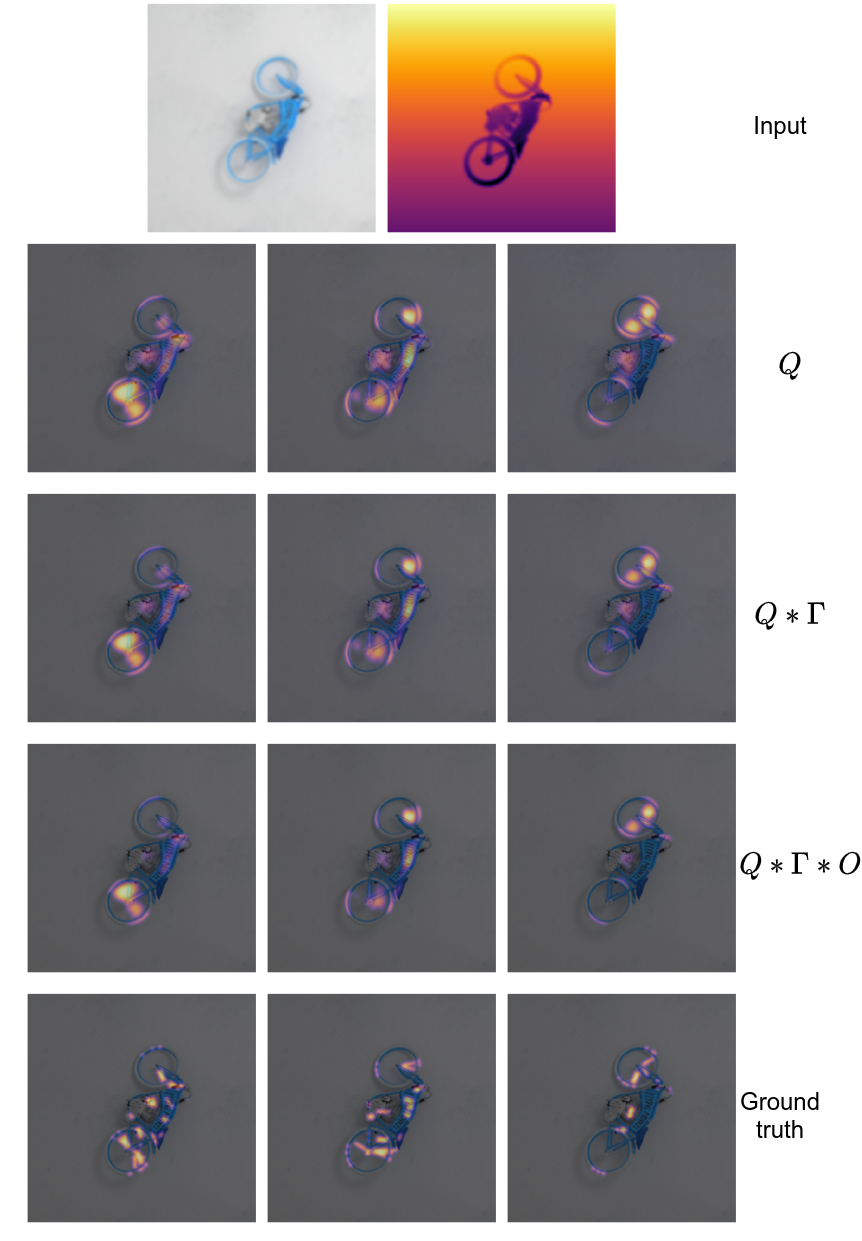}
    \caption{Intermediate results when reconstructing the quality map. $Q$ alone is noisy as the output of a regression problem. Multiplication by $\Gamma$ smooths the quality map pixel-wise, while $O$ filters outliers bin-wise. The final estimation is much clearer and precise.}
    \label{fig:results}
    \vspace{-0.5cm}
\end{figure}

\noindent\textit{\underline{Training}:} Each map is separately supervised: we minimize the Mean Square Error (MSE) of the real-valued $Q,\mbox{ } cos(2\Phi),\mbox{ } sin(2\Phi)$ and $\Omega$ and their respective GTs, and we force a Binary Cross-Entropy loss (BCE) on $O$ and $\Gamma$. Next, we employ an attentive loss that directly minimizes the MSE between $Q * O$ (element-wise multiplication) and the GT quality map. This attention mechanism drives the network's focus over regions of the feature map that correspond to filled bins and thus regions nearby a valid grasp center. We found it useful to scale the MSE losses by multiplying them with the number of bins $N$. The total objective function takes the form:
\begin{align}
\pazocal{L} ={} & \pazocal{L}_{BCE}(O) + \pazocal{L}_{BCE}(\Gamma) \nonumber \\
      & + N * \{\pazocal{L}_{MSE}(Q) + \pazocal{L}_{MSE}(cos(2\Phi)) + \pazocal{L}_{MSE}(sin(2\Phi)) \nonumber \\
      & + \pazocal{L}_{MSE}(\Omega) + \pazocal{L}_{MSE}(Q * O)\}
\end{align}
\noindent\textit{\underline{Inference}:} First, $Q$ and $\Gamma$ are multiplied to obtain a graspability-refined quality map. This can be viewed as a pixel-wise prior regularization, where $\Gamma$ is the prior probability of a pixel to be a grasping point and $Q$ is the posterior, measuring its grasping quality. This product is multiplied by $O$ to filter out values in empty bins, resulting in the final quality map, $Q*\Gamma*O$. Fig~\ref{fig:results} shows the intermediate effects of the quality map refinement on real images. Finally, we choose the optimum grasping center as the global maximum of the quality map and retrieve the respective values of $\Phi$ and $\Omega$ to reconstruct a grasping box.

\noindent\textit{\underline{Model zoo}:}
We embed ORANGE to two off-the-shelf architectures, GGCNN2 \cite{GGCNN_IJRR} and the larger U-Net \cite{Unet15}, both able of performing segmentation. While these models have totally different capacity, we show that both can perform significantly better when trained with ORANGE.

\section{Experiments \& Discussion}
We evaluate ORANGE on Jacquard and Cornell, following the standard $90$/$10$\% split for training and testing respectively, and without any data augmentation. The input depth images are resized from their initial size to $320 \times 320$, to allow for higher training speed. We train the whole network end-to-end employing an early stopping strategy with an initial learning rate of $0.002$ that decays exponentially. On Cornell, we warm-start the network with weights trained on Jacquard and then finetune without any data augmentation. Following prior literature, we adopt IoU@$0.25$ and $0.30$, as our evaluation metrics. Finally, we provide results for the Jacquard virtual grasping evaluation and for a real-robot grasp experiment with household objects, previously unseen during training, showcasing the benefit of planning grasps from the disentangled grasp map predictions.

\subsection{Ablation study}
We inspect the different combinations of the individual components of ORANGE in Table \ref{tab:ablation}, employing U-Net as the base model, as it has more capacity to absorb the multiple grasp representations. Our full proposed model, with the pixel-wise graspability and the bin-wise orientation attention, performs better when using 3 bins compared to 6 for the lower threshold; discretizing the angle space into N bins means N regressions for the model to learn and N classes to identify. In particular, for the angle range of $\{-\pi/2,\pi/2 \}$ in the antipodal grasps, the $N=6$ discretization, divides into bins of $30^o$ range, i.e. there are smaller differences in the appearances among neighboring orientations, and it requires more regressions, making it more difficult to disentangle the multiple grasping orientations. 

\begin{table*}
\vspace{-0.1cm}
    \centering
    \begin{adjustbox}{width=\linewidth}
    \begin{tabular}{c|c|c|c|c|c|c|c|c|c||c|c}
    \toprule                  
    \multicolumn{10}{c}{Design Choices} &  \multicolumn{2}{c}{Accuracy Threshold} \\ \hline
         Base network & regression & graspability & bin class. & attention & binary map & max jaw size & min jaw size & $N=3$ & $N=6$ & 0.25 & 0.30 \\ \hline
         \midrule
         U-Net & \color{orange}\checkmark & \color{orange}\checkmark &  \color{orange}\checkmark & \color{orange}\checkmark&   &   &   \color{orange}\checkmark &  \color{orange}\checkmark &  & \color{orange}\textbf{94.71} & \color{orange}\textbf{92.65} \\ \hline
               & \checkmark & \checkmark &  \checkmark & \checkmark&   &   &   \checkmark &             & \checkmark & 91.51 & 89.07 \\ \hline
               & \checkmark &            &  \checkmark & \checkmark&   &   &   \checkmark &  \checkmark &  & 92.34 & 90.44\\ \hline
               & \checkmark & \checkmark &  \checkmark &           &   &   &   \checkmark &  \checkmark &  & 93.36  & 90.90 \\ \hline
               & \checkmark & \checkmark &  \checkmark & \checkmark&           &\checkmark&             & \checkmark &  & 94.11 & 91.83\\ \hline
               & \checkmark & \checkmark &  \checkmark & \checkmark& \checkmark &   &   \checkmark &  \checkmark &  & 91.75 & 90.27 \\ \hline
               & \checkmark &            &             &           & \checkmark  &   &       &      &  & 89.85  & 88.13\\ \hline

         \midrule
         GGCNN2 & \color{orange}\checkmark & \color{orange}\checkmark &  \color{orange}\checkmark & \color{orange}\checkmark&   &   &   \color{orange}\checkmark &  \color{orange}\checkmark &  &\color{orange} \textbf{88.92} & \color{orange}\textbf{85.94} \\ \hline
               & \checkmark &  &  \checkmark & \checkmark&   &   &   \checkmark &  \checkmark &  & 87.88  & 85.52\\ \hline
               & \checkmark &            &             &           & \checkmark  &   &       &      &  & 85.23 & 82.67 \\ \hline    
               \bottomrule
               \end{tabular}
\end{adjustbox}
    \caption{Ablation study over different design choices for both ORANGE implementations with U-Net and GGCNN2. For each instantiation, the accuracy (\%) is reported over different thresholds of the IoU index. With \color{orange}{orange} \color{black} we denote the full proposed architecture.}
    \label{tab:ablation}
        \vspace{-0.5cm}
\end{table*}

The application of the pixel-wise graspability $\Gamma$ on the quality maps $Q$ has an evident benefit on the model, as the graspability loss focuses locally on the best grasp points. The jaw size selection during the construction of the GT maps also affects the performance of ORANGE. As we discussed in Sec. \ref{sec:dataset_discussion}, the jaw size is a feature indifferent to the object. We experimented with the minimum or maximum possible sizes. Intuitively, using the minimum jaw size produces boxes closer to the boundaries of the object, and thus can be easier to be segmented by U-Net, as seen in the table. An important decision is the use of binary quality maps in the GT data synthesis \cite{GGCNN_IJRR}. Using binary maps in ORANGE produces $4$\% less accurate (IoU@$0.25$) grasp predictions compared to our approach. This greedy solution creates higher confusion about which pixel is the grasp point. ORANGE seems to mitigate this confusion achieving an accuracy of $91.75$\%, while a U-Net implemented as suggested in \cite{GGCNN_IJRR}, succeeds an $89.85$\% at IoU$0.25$. 
Lastly, we also improve GGCNN2 from $85.23$\% of the original implementation into $88.92$\%, confirming ORANGE's model-agnostic character.

\subsection{Comparisons to prior literature}
We compare our best configuration to all known works for Jacquard in Table \ref{tab:comparative_results1}. Note that most related works predict grasps from multi-modal data (i.e. RGB, RGB-D, or RGD). Nevertheless, our pure depth-based ORANGE outperforms all existing approaches on Jacquard to achieve a new state-of-the-art of $94.7$\% IoU@$0.25$. Note that we are also slightly better than the RGBD method of the concurrent \cite{kumra2019antipodal} and by $1\%$ better than their depth result. We attribute this improvement to two factors: (i) the augmented grasp map representation, and (ii) the bin-wise attention of the orientation estimation over the quality maps, which disentangles the overlaps of multiple orientations per grasp point. We expect even better performance if we also use the RGB channels, however, this is beyond the scope of our work that focuses on proposing a novel grasp map representation that disentangles grasp orientations for effective grasp planning.

Subsequently, we evaluate ORANGE on Cornell. This dataset is smaller and contains manual and noisy annotations. Prior literature has used massive data augmentation to achieve good learning performance. However, such results do not scale easily to the real-world. We, on the other hand, evaluate the transferability of our network to a new dataset. To achieve this, we employ the technique of ``warm-starting'' \cite{KirkpatrickPRVD16}, a method used for initializing the weights of a pretrained network, while not allowing for catastrophic forgetting of the previously learned representations. We train only on 90\% of the Cornell dataset, without using any data augmentation, contrary to all related works, and achieve a high testing accuracy.

Table \ref{tab:comparative_results2} presents the comparative results for Cornell. Specifically, the depth-based ORANGE approach achieves $91.1$\% with U-Net and $87.5$\% with GGCNN2. This accuracy score, although not state-of-the-art, is still a very high performance among the depth-based methods, e.g. GGCNN2 achieves $78.6$\% when train/tested (no data-augmentation) on Cornell, which ORANGE improves by $9$\%. It is natural to expect improved performance with data augmentation, yet this is not reported in this work, as our main focus is to stress the benefit of ORANGE and its novel grasp map representation that effortlessly generalizes to new data.

\begin{table}
    \centering
    \adjustbox{width=0.48\textwidth}{
{\footnotesize 
    \begin{tabular}{l|c|c}
    \toprule
        methods & modality & Accuracy (\%) \\ \hline
        \midrule
        Morrison et al. \cite{GGCNN_IJRR} & D & 85.2 \\ \hline
        Depierre et al. \cite{Jacquard} &  RGB-D & 74.2   \\ \hline
        Zhou et al. \cite{Zhou2018} & RGB & 91.8 \\ \hline
        Zhou et al. \cite{Zhou2018} & RGD & 92.8 \\ \hline
        Zhang et al. \cite{Zhang} & RGB &  90.4 \\ \hline
        Zhang et al. \cite{Zhang} & RGD & 93.6 \\ \hline
        Kumra et al. \cite{kumra2019antipodal}& D & 93.7 \\ \hline
        Kumra et al. \cite{kumra2019antipodal}& RGB & 91.8\\ \hline
        Kumra et al. \cite{kumra2019antipodal}& RGBD & 94.6\\ \hline
        ORANGE with GGCNN2 (ours)& D & \textbf{88.9} \\ \hline
        ORANGE with U-Net (ours) & D & \textbf{94.7} \\ \hline
        \bottomrule
    \end{tabular}  }}
    \caption{Comparative results for the Jacquard dataset.}
    \label{tab:comparative_results1}

   \adjustbox{width=0.48\textwidth}{
{\footnotesize
    \begin{tabular}{l|c|c}
    \toprule
        methods & modality & Accuracy (\%) \\ \hline
        \midrule
        Morrison et al. \cite{GGCNN_IJRR} & D & 78.6 \\ \hline
        Depierre et al. \cite{Jacquard} trained on Jacquard &  RGB-D & 81.92 \\ \hline
        Depierre et al. \cite{Jacquard} trained on Cornell &  RGB-D & 86.88   \\ \hline
        Zhou et al. \cite{Zhou2018} & RGB & 97.7 \\ \hline
        Zhang et al. \cite{Zhang} & RGB &  93.6 \\ \hline
        Zhang et al. \cite{Zhang} & RGD & 92.3 \\ \hline
        Guo et al. \cite{Guo2017} & RGB & 93.2 \\ \hline
        Chu et al. \cite{Chu2018} & RGB & 94.4 \\ \hline
        Chu et al. \cite{Chu2018} & RGB-D & 96.0 \\ \hline
        Wang et al. \cite{Wang} & RGB-D & 94.4 \\ \hline
        Kumra et al. \cite{kumra2019antipodal}& D& 94.3\\ \hline
        Kumra et al. \cite{kumra2019antipodal}& RGB & 95.5 \\ \hline
        Kumra et al. \cite{kumra2019antipodal}& RGBD & \textit{96.6} \\ \hline
        ORANGE with GGCNN2 (ours)& D & \textbf{87.5} \\ \hline
        ORANGE with U-Net (ours) & D & \textbf{91.1} \\ \hline
        \bottomrule
    \end{tabular}}}
    \caption{Comparative results for the Cornell dataset. ORANGE was trained on Cornell with weights `warm-started' from the Jacquard dataset, but without using additional data augmentation. All other works, unless stated otherwise, were trained/tested on Cornell using data augmentation.}
     \label{tab:comparative_results2}

    \centering
      {\footnotesize
    \begin{tabular}{l|c|c}
    \toprule
        object & gripper & hand  \\ \hline
        \midrule
       Chips Can (\#1) & 4/5 & 0/5 \\ \hline
       Cracker Box (\#3) & 5/5 & 4/5\\ \hline
       Pear (\#16) & 4/5 & 2/5\\ \hline
       Mug (\#25) & 3/5 & 4/5 \\ \hline
       Foam brick (\#61) & 5/5 & 5/5 \\ \hline
       Total & 21/25   & 15/25  \\
        \bottomrule
    \end{tabular}}
    \caption{\small Real robot experimental results with a bi-manual robot (Fig.~\ref{fig:robot_motivation}, equipped with a gripper and a five-finger under-actuated hand. We report grasp success across five different YCB objects \cite{YCB}, for five grasp trials per experiment, using our best scoring ORANGE model for grasp generation.}
     \label{tab:robot_results}
     \vspace{-0.5cm}
\end{table}

\subsection{Evaluation of robot grasp synthesis}

\noindent\textit{Simulated tests.}
We evaluate a test-set of grasps on the Jacquard online simulator using the Simulated Grasp Test (SGT) \cite{Jacquard}. In SGT, a grasp is successful if the object is lifted, moved away, and dropped at a given location by the simulated robot. We randomly select a set of $500$ images from the test-set of Jacquard and we compare the proposed full model of ORANGE with UNet and GGCNN2 backbone, which result in $88.2\%$ and $85\%$ grasp success, respectively. Notably, the original GGCNN2\cite{GGCNN_IJRR} achieves $75.6\%$, and the UNet implemented with the grasp representation of GGCNN2 $80\%$. The simulated grasping results confirm that the ORANGE framework can generate more reliable grasp representations leading to a higher success rate in the grasp execution.

\noindent\textit{Real robot evaluation.}
To demonstrate the effectiveness of ORANGE, with our best scoring model (Table \ref{tab:ablation}), we conduct experiments with the bi-manual mobile manipulator robot TIAGo++, equipped with one gripper and a five-fingered underactuated hand. We leverage our robot's properties to study how the different orientations in the grasp maps can enable a successful robot grasp\footnote{Code available: \url{https://github.com/gchal/tiago_grasp_plan}}.

For this experiment, we chose a set of five objects from the YCB object set \cite{YCB}, for which we conduct $10$ grasps per item; $5$ for the left and $5$ for the right arm (with a gripper and a hand end-effector, respectively). We place the robot in front of a table and capture the object depth image from the robot's built-in camera. Note that this experiment is more challenging both because the camera viewpoint is much different than the training dataset setting, as well as compared to other related robotic experiments, that install a static camera facing the table vertically \cite{kumra2019antipodal} and plan planar pinching grasps \cite{morrison2018closing}. We, on the other hand, plan open-loop collision-free trajectories \cite{sucan2012the-open-motion-planning-library} considering the robot's arm and torso motion for planning the trajectory towards a generated target grasp vector by ORANGE. A grasp is successful when the robot holds the object in the air for $10$ seconds. Note that for our experiment, we collect the best grasp point across all bins (i.e., for all predicted orientations) and attempt the ones that are within the feasibility set of the workspace of each arm to showcase the importance of parsing the possible grasp angles.

\begin{figure}
\includegraphics[width=0.48\textwidth]{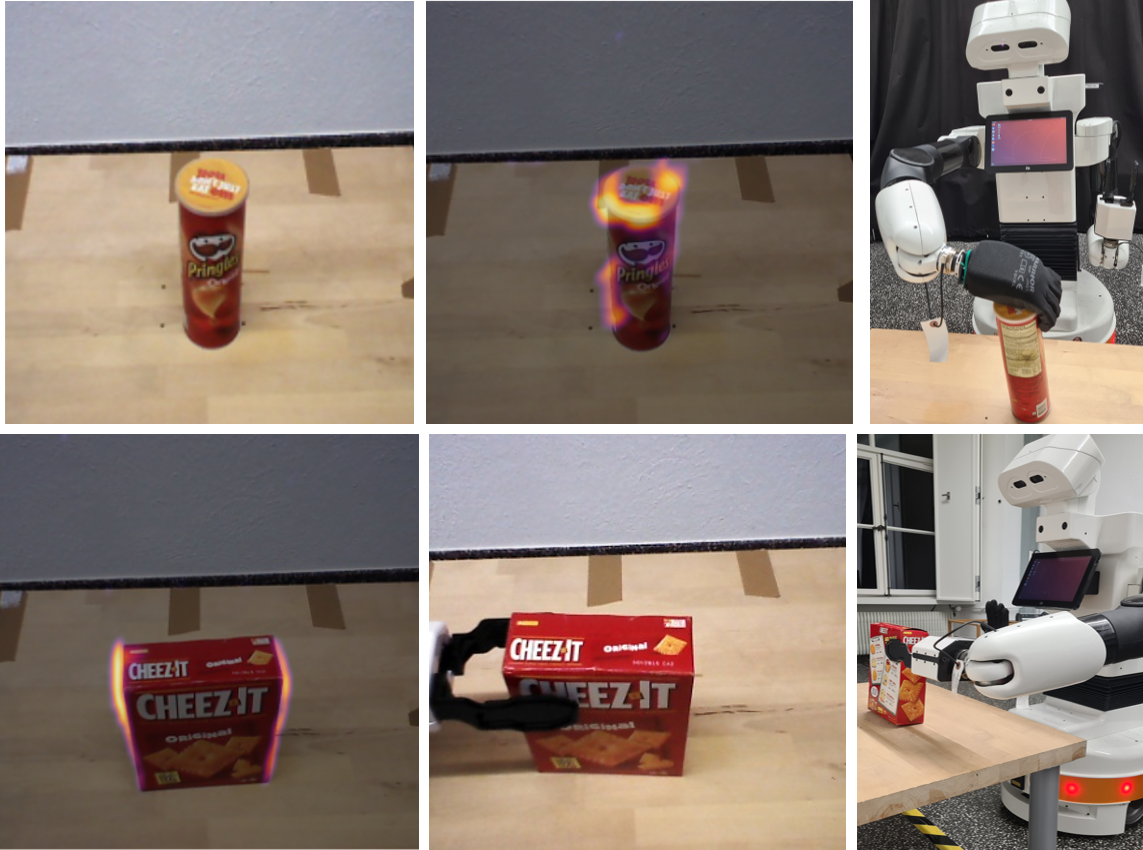}
 \caption{\textbf{Upper:} Object chips can (\#1) in the robot's field of view. The colormap (middle) shows the grasp map for a selected orientation bin. The robot attempts a grasp of the object at the predicted grasp point.
 \textbf{Lower:} Object cracker box (\#3) with the grasp map along the selected orientation bin. Robot grasp the objected at the best grasp position of the bin, and successfully lifts the object.} \label{fig:tiago_grasp}
 \vspace{-0.5cm}
\end{figure}

Table \ref{tab:robot_results} summarizes the results of the robotic experiment. While with the gripper we are able to grasp most objects ($84\%$ grasp success), grasping with the robotic hand is more challenging. As we can see in Fig. \ref{fig:tiago_grasp} for the Chips Can, ORANGE delivers a very good grasp map and the robot reaches for the targeted grasp point; however, it is unable to lift the object in the air, due to the morphology of the hand and other parameters, e.g. low friction between object and hand. While using the gripper we are able to achieve good grasps in feasible positions for the robot's left arm (Fig. \ref{fig:tiago_grasp}-lower), we sometimes fail when it comes to objects like the mug, for which the predicted grasps are focusing on the handle. Interestingly, we are able to grasp the mug with the hand, as this grasp requires finer manipulation (Fig. \ref{fig:robot_motivation}). The results of this experiment highlight two findings: (i) the advantage of acquiring a disentanglement of the potential grasp orientations provides a promising framework for planning feasible robot grasps, especially with bi-manual and mobile manipulator robots. A possible future research direction concerns the learning of a policy for selecting the grasp points per orientation; (ii) a good visual grasp generator can only be a good indicator for a successful grasp. We believe that a combination of the effectiveness of ORANGE fused with tactile feedback can potentially provide a more powerful tool for effective grasping. 

\section{Conclusions \& Future work}
In this paper, we discussed and addressed the problem of multiple grasping orientations in objects that confuse recent visual regressors. To alleviate this challenge, we proposed a novel view on the grasp representation, allowing the co-occurrence of multiple grasp orientations. On top of that, we introduced ORANGE, a model-agnostic approach that jointly solves an angle-bin classification and real-value angle regression while exploiting the former to guide a graspability attention mechanism over the grasp quality map. Extensive experimental results justified the effectiveness of ORANGE design, which achieves state-of-the-art performance using only the depth modality. Our robotic experiment showed the increased benefit of planning grasps according to the orientations that are feasible in the robot's workspace. An interesting future direction is to jointly reason about the objects' grasping points, shape, and category. The quality of the generated grasps can also be ranked in an adversarial setting while interacting with real objects for learning to identify task-related grasp points.

\clearpage
\bibliographystyle{IEEEtran}

\bibliography{Grasp_references.bib}
\balance
\end{document}